\documentclass[letterpaper,10pt,conference]{ieeeconf}
\IEEEoverridecommandlockouts
\usepackage{amsmath,amssymb,graphicx,booktabs,url,tikz,array,colortbl,needspace}
\usepackage[hidelinks]{hyperref}
\definecolor{RiseBlue}{HTML}{397DB7}
\definecolor{SettleGreen}{HTML}{398B66}
\definecolor{SupportOrange}{HTML}{DD841C}
\definecolor{LoadPurple}{HTML}{79638F}

\usepackage{xcolor}

\makeatletter
\def\@IEEEauthorblockAtopspace{8pt}
\let\ROOForiginalmakecaption\@makecaption
\long\def\@makecaption#1#2{%
  \ifx\@captype\@IEEEtablestring
    \par\begingroup\footnotesize\normalfont
    \noindent Table~\thetable: #2\par
    \endgroup\@IEEEtablecaptionsepspace
  \else
    \ROOForiginalmakecaption{#1}{#2}%
  \fi}

\makeatother
\title{\LARGE\bf ROOF: Learning Humanoid Fall Recovery under Overhead Constraints}
\author{\authorblockN{Yidan Lu$^{1}$, Jiayuan Yu$^{2}$, Zongrui Cai$^{1}$, Jingwen Sun$^{1}$, Jinrui Han$^{1}$,\\
Jing Wang$^{1}$, Ping Luo$^{1}$, Peng Lu$^{1,\dagger}$\thanks{$^{\dagger}$Corresponding author: \href{mailto:lupeng@hku.hk}{lupeng@hku.hk}. \href{mailto:ydlu@connect.hku.hk}{ydlu@connect.hku.hk}.}}
\authorblockA{$^{1}$The University of Hong Kong \qquad $^{2}$Chinese Academy of Sciences}}
\IEEEaftertitletext{\vspace{-12pt}\begin{center}
\offinterlineskip
\definecolor{SupineBand}{HTML}{88A6C4}
\definecolor{ProneBand}{HTML}{B2C9A3}
\setlength{\fboxsep}{0pt}
\vbox{
\hbox{\includegraphics[width=\dimexpr\textwidth/5\relax]{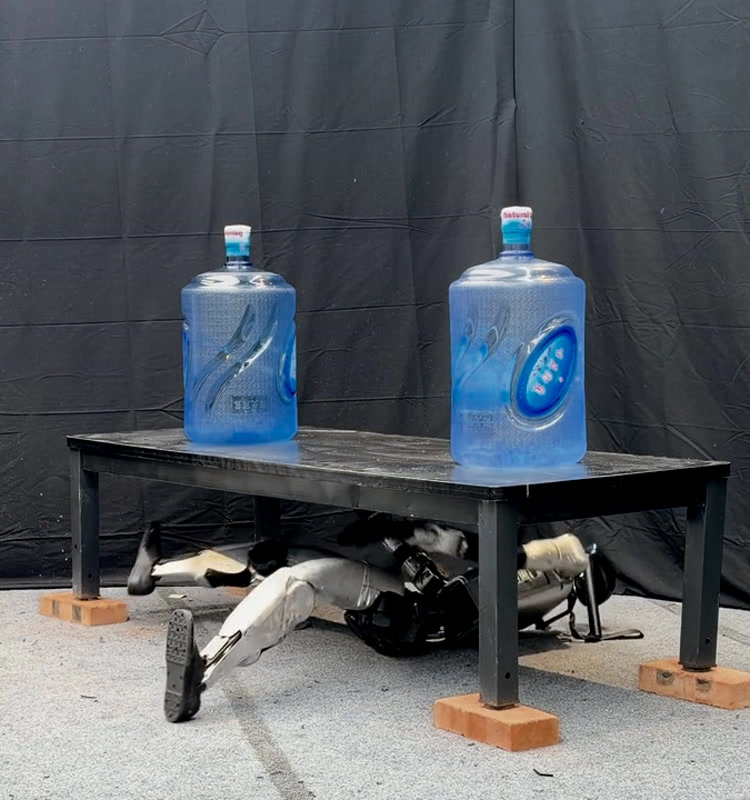}%
\includegraphics[width=\dimexpr\textwidth/5\relax]{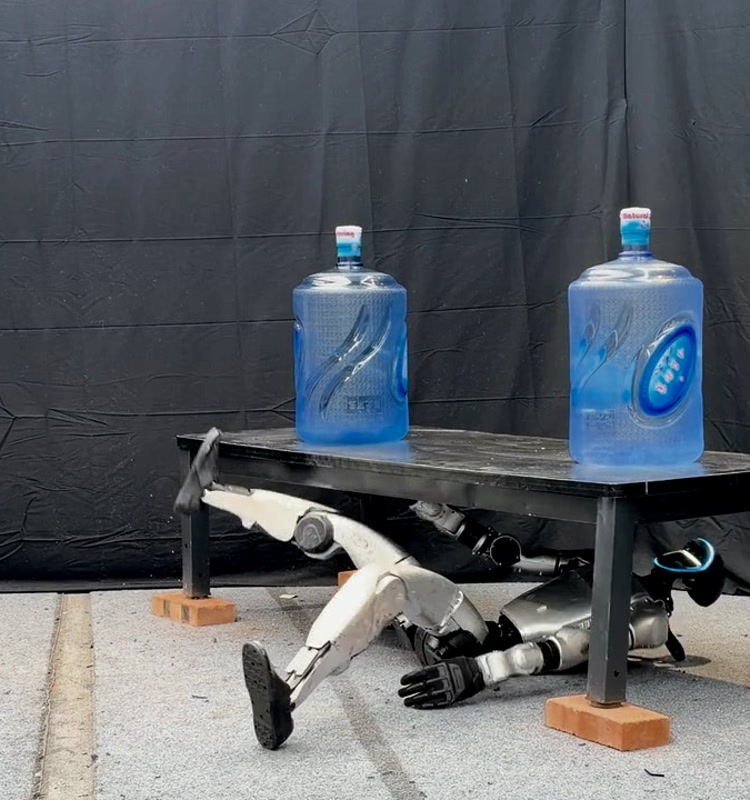}%
\includegraphics[width=\dimexpr\textwidth/5\relax]{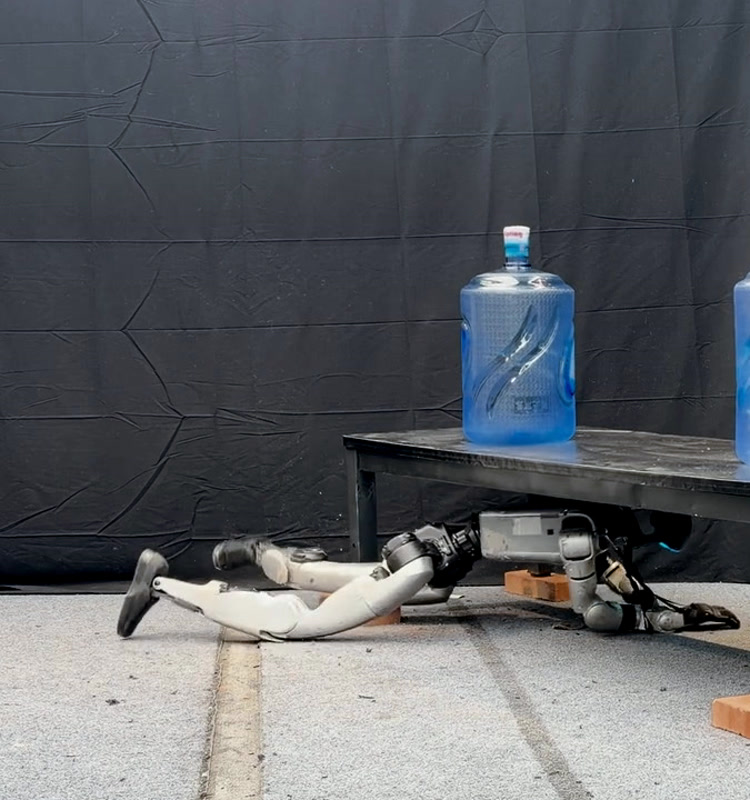}%
\includegraphics[width=\dimexpr\textwidth/5\relax]{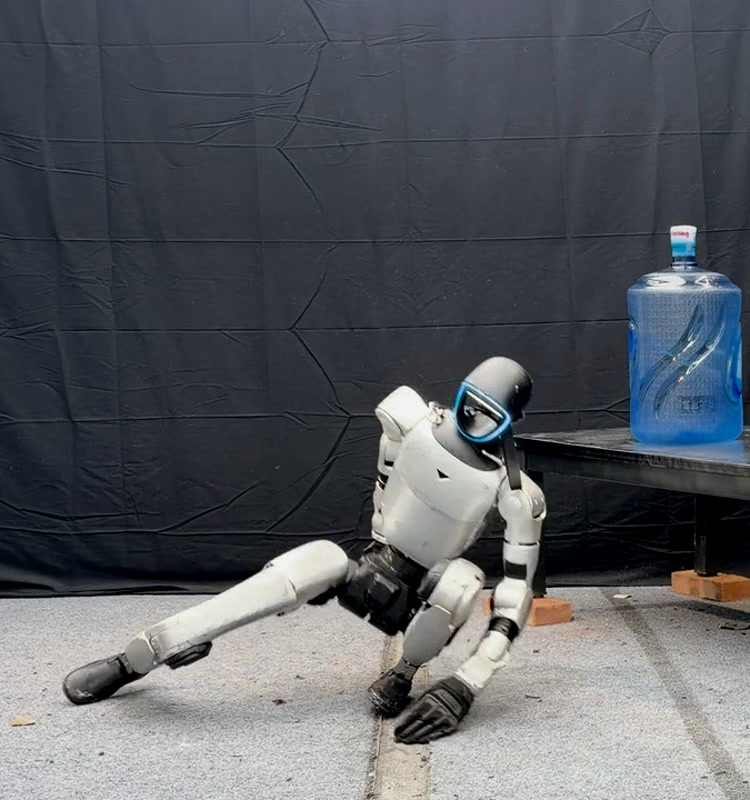}%
\includegraphics[width=\dimexpr\textwidth/5\relax]{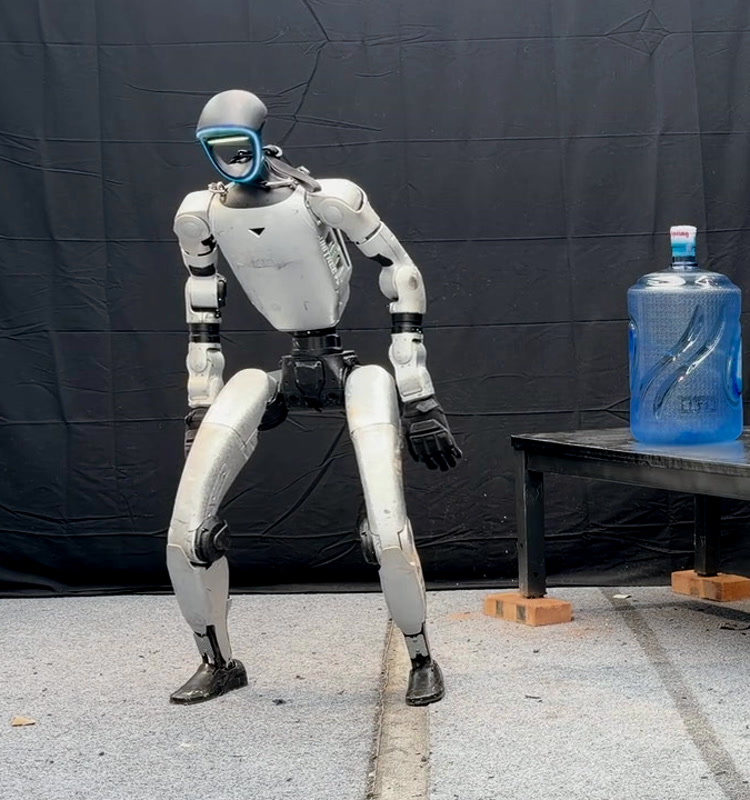}}%
\hbox{\colorbox{SupineBand}{\parbox[c][10pt][c]{\textwidth}{\centering\sffamily\bfseries\footnotesize Recovery Beneath a Table}}}%
\hbox{\includegraphics[width=\dimexpr\textwidth/5\relax]{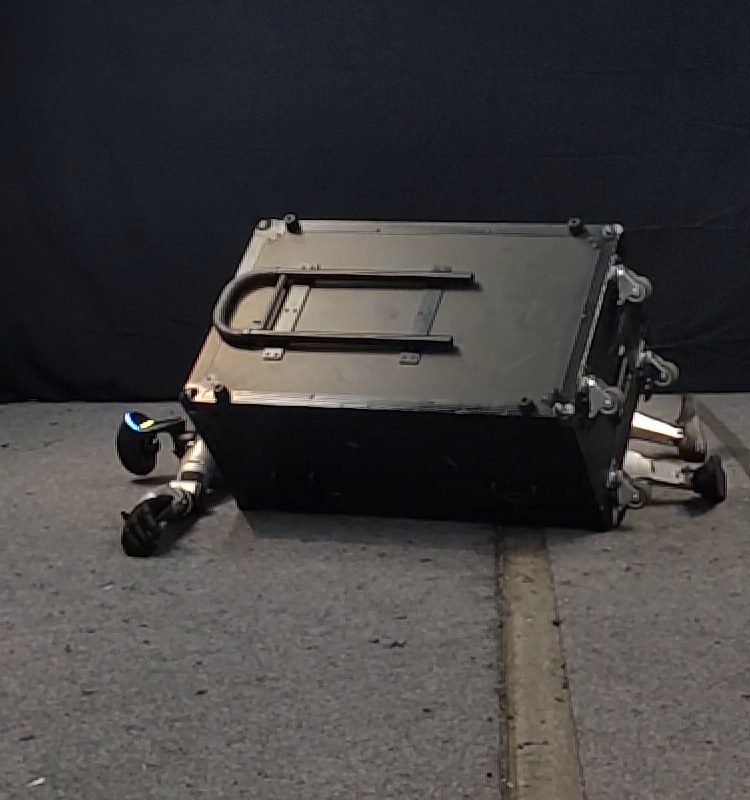}%
\includegraphics[width=\dimexpr\textwidth/5\relax]{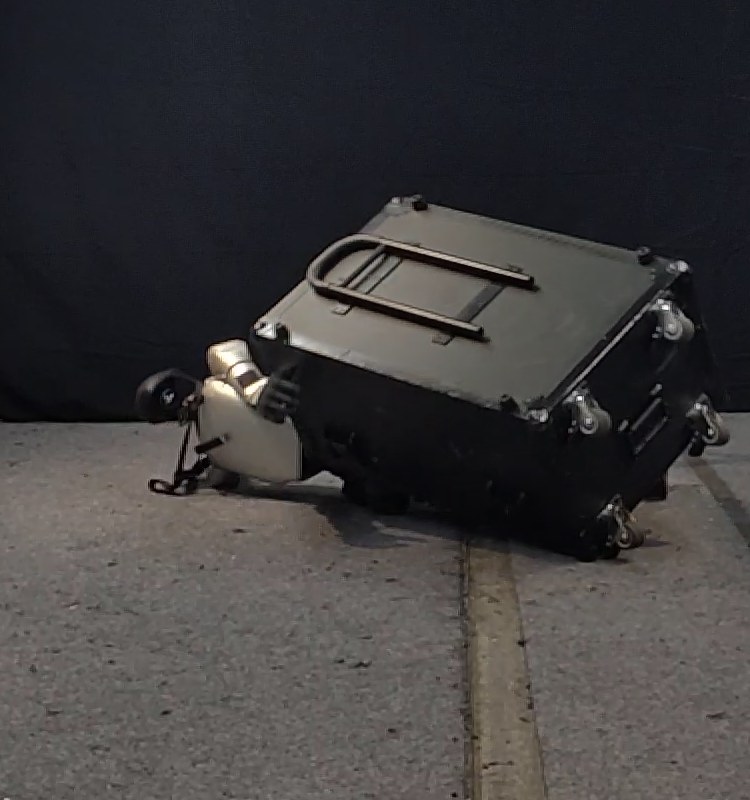}%
\includegraphics[width=\dimexpr\textwidth/5\relax]{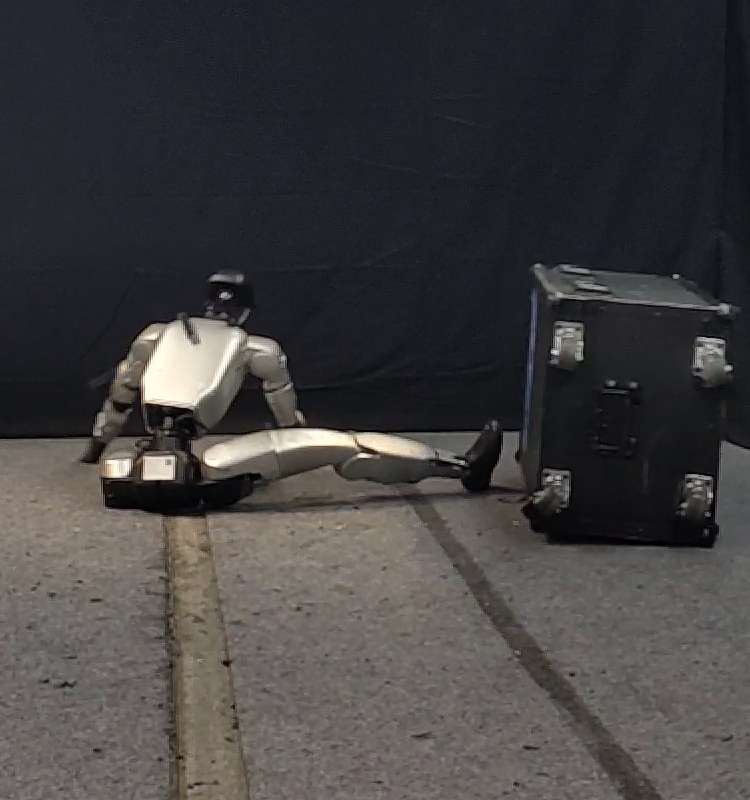}%
\includegraphics[width=\dimexpr\textwidth/5\relax]{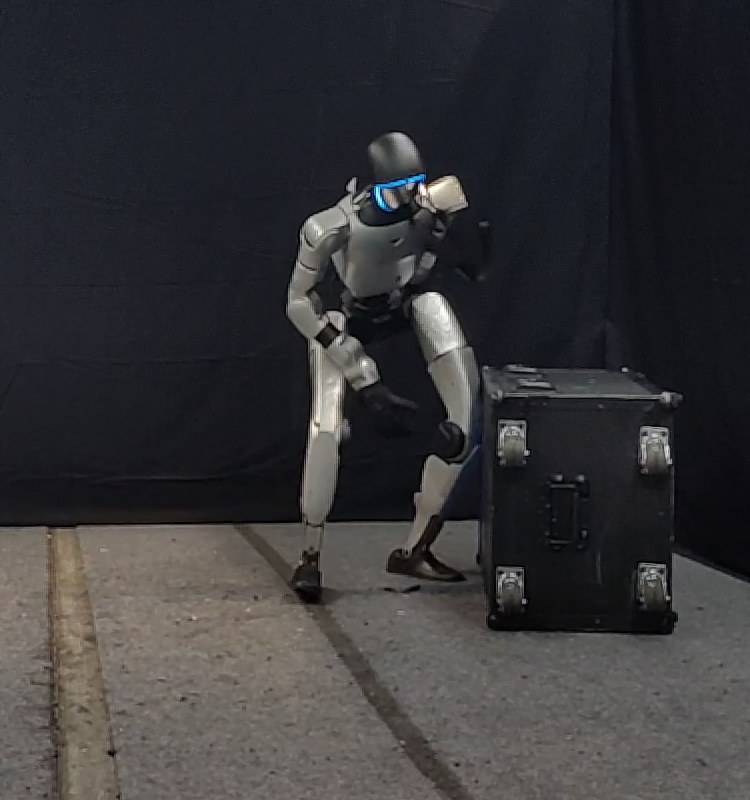}%
\includegraphics[width=\dimexpr\textwidth/5\relax]{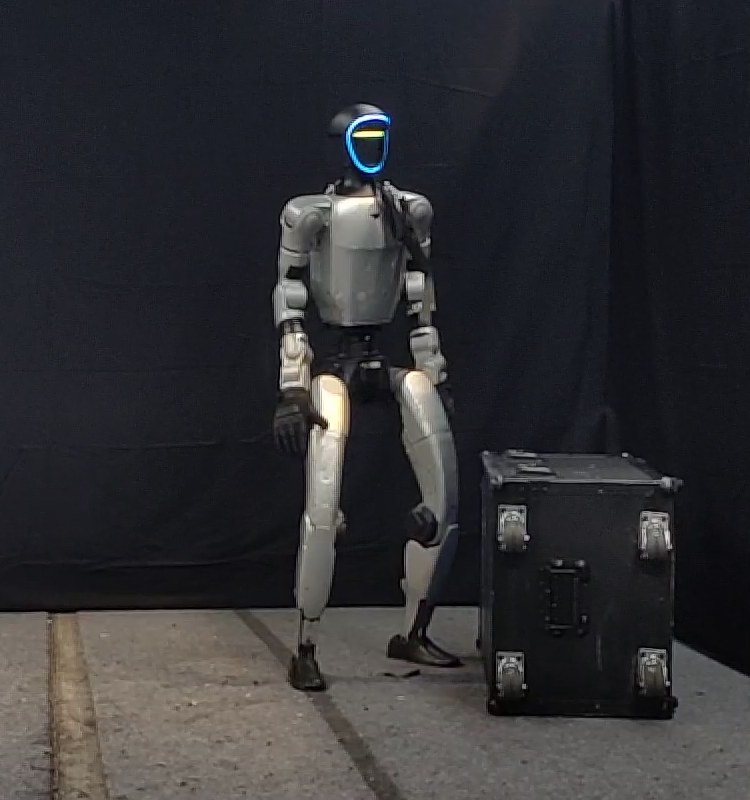}}%
\hbox{\colorbox{ProneBand}{\parbox[c][10pt][c]{\textwidth}{\centering\sffamily\bfseries\footnotesize Recovery Under a Rigid Equipment Case}}}%
}
\par\vspace{4pt}
\begin{minipage}{.98\textwidth}\footnotesize
Figure 1. Recovery under overhead constraints. Beneath a table (top), the robot adjusts its support and clears the table edge before standing. Beneath a rigid equipment case (bottom), interaction tips the obstacle aside and permits the rise. Each row shows successive frames from one continuous recovery test on a Unitree G1.
\end{minipage}
\end{center}
}
\begin{document}
\maketitle
\refstepcounter{figure}\label{fig:teaser}
\thispagestyle{empty}\pagestyle{empty}
\begin{abstract}

A fallen humanoid beneath an overhead obstacle faces a coupled recovery problem.
It must create clearance while organizing support for a stable rise, since attempting to stand before making room can further increase obstruction.
We introduce a reinforcement learning method with contact-supported geometric escape guidance, which rewards incremental reduction of robot-obstacle overlap without prescribing an escape path.
To achieve this, a learned motion prior, together with posture-dependent support and joint-stall costs, is proposed to shape a single policy that learns when to clear the obstacle, how to establish support, and when to complete the rise, all using proprioception alone.
In simulation, it recovers in 88.5\% of trials beneath plates that can only be lifted, compared with 18.9\% without geometric guidance, and in 98.6\% of trials beneath free plates with randomized size, mass and placement. 
On a Unitree G1, the deployed policy recovered in 95 of 150 attempts across seven real-world obstacle scenes, including furniture and multi-object arrangements absent from training.
Project page: \href{https://lu-yidan.github.io/ROOF/}{\textcolor{blue}{\nolinkurl{https://lu-yidan.github.io/ROOF/}}}.

\end{abstract}
\section{Introduction}

A humanoid that falls beneath a table, shelf or fallen board cannot necessarily use its usual get-up motion. 
Raising the torso might drive it into the obstacle, while remaining in place leaves too little room to stand. 
Recovery, therefore, requires the robot to change the configuration that blocks its rise, where it may move out from beneath the obstacle, push a movable obstacle aside, or combine both actions. 
It must do so while preserving or establishing support for the subsequent rise. 
Figure~\ref{fig:teaser} shows examples of two such recoveries on a Unitree G1.

Learned controllers have made humanoid recovery increasingly robust across fallen postures and terrain, with the recent integration of fall mitigation, locomotion and terrain-aware behaviors~\cite{humanup,host,firm,vigor,unirelo}.
These methods address how a robot can rise from a difficult configuration. 
Recovery beneath an overhead obstacle poses a different challenge, where the configuration itself may not permit rising. 
The robot must first create clearance through interaction with its surroundings, then use the resulting space to stand. 
In this case, the required action is scene-dependent.
For example, a movable board may yield when pushed, while a fixed table requires the robot to move beyond its edge. 
A method that encodes a specific escape direction or contact sequence would need to be redesigned for each scene.

We formulate this problem as learning a single recovery behavior that couples clearance creation with rising.
The key challenge is to provide useful guidance about whether the robot is making room without specifying how it must do so. 
We address it with a contact-supported geometric reward that measures progress in reducing robot-obstacle overlap and increasing separation. 
The reward encourages progress while the robot remains low and supported, yet leaves the escape direction and choice of contact to the policy. 
A learned human-motion prior encourages coordinated movement, while posture-dependent recovery objectives and costs shape the transition from supported repositioning to a controlled rise. 
Obstacle geometry is available during training to compute this guidance; at deployment, the policy uses proprioception alone.

Training proceeds from the flat-ground recovery to obstacle interaction, retaining ordinary recovery episodes as the policy learns to interact with obstacles. 
We evaluate the method in simulation with both guided plates, which can only be lifted, and free plates with randomized size, mass and placement. 
We then test the learned policy on a physical Unitree G1 beneath obstacles and arrangements absent from training. 
The policy recovers in $88.5\%$ of guided-plate simulation trials, compared with $18.9\%$ without geometric guidance, and in $98.6\%$ of free-plate trials. 
On hardware, it recovers in 95 of 150 attempts across seven obstacle scenes, including furniture and multi-object arrangements.
Together, our results show why recovery under overhead constraints must be evaluated as a sequence of creating room and standing, rather than by uprightness alone.

Our contributions are:
\begin{itemize}
    \item We formulate humanoid recovery beneath overhead obstacles as a coupled problem of creating clearance and achieving a supported rise, and learn both behaviors with one proprioceptive policy.
    \item We introduce contact-supported geometric reward guidance that measures progress in reducing robot-obstacle overlap without specifying an escape direction or contact sequence. 
    Combined with a learned motion prior, it supports both coordinated escape and standing.
    \item We evaluate the method in simulation with movable obstacles and on a Unitree G1 across seven real-world scenes, demonstrating recovery under furniture and multi-object arrangements absent from training.
\end{itemize}

\section{Related Work}
\subsection{Learning Coordinated Humanoid Recovery}
Humanoid stand-up has progressed from finding feasible motions to making recovery robust and deployable. Morimoto and Doya demonstrated hierarchical reinforcement learning for stand-up behavior on a real robot~\cite{morimoto}. Later policies learned quadruped recovery~\cite{hwangbo} and integrated rapid recovery with locomotion and soccer skills~\cite{soccer}.

For larger humanoids, HumanUP separates motion discovery from refinement into deployable trajectories across postures and terrains~\cite{humanup}. HoST combines posture-adaptive learning with an assistance curriculum and motion constraints~\cite{host}. FIRM extends recovery to a unified fall-safety policy, using demonstrations and diffusion-based memory to connect fall mitigation with standing up~\cite{firm}. These advances address diverse postures and fall conditions. We study the additional challenge of creating room beneath an overhead obstacle while preserving the support needed to rise.

Motion priors encourage coordinated behavior while task rewards specify what the robot must accomplish. DeepMimic combines reference tracking with task objectives~\cite{deepmimic}; GAIL and AMP learn behavioral rewards from examples without requiring a particular reference trajectory~\cite{gail,amp,ampreal}. ASE and CALM learn skill representations reusable across tasks~\cite{ase,calm}. State-dependent AMP combines locomotion and recovery~\cite{stateamp}, while StableMimic and Heracles connect motion tracking with recovery generation~\cite{stablemimic,heracles}.

Diffusion models can represent action policies directly~\cite{diffusionpolicy} or provide a frozen motion reward, as in SMP~\cite{smp}. We use SMP for movement coordination and discard the prior model at deployment. A general motion prior does not specify how to clear a particular obstacle; geometric guidance supplies this task-specific feedback.

A successful simulated recovery can still be oscillatory or sensitive to dynamics. Dynamics randomization addresses model mismatch~\cite{dynrand}, while Lipschitz-constrained policies regularize humanoid action sensitivity~\cite{lcp}. Recovery methods also limit speed and refine discovered motions~\cite{host,humanup}. Our refinement similarly addresses motion quality, with foot-motion and joint-stall costs tailored to support transfer and obstructed movement.

\subsection{Contact-Aware Recovery and Interaction}
Environmental contact can both assist recovery and constrain it. Contact-planning methods redirect falling momentum through multiple contacts~\cite{multicontactfalls}, jointly optimize stabilization and support changes~\cite{contacttree}, and incorporate sliding and detaching contacts into stand-up motions~\cite{slidecontact}. Learning-based methods also exploit non-foot contacts: arm contact can reduce fall damage and assist recovery in a legged manipulator~\cite{armrecovery}.

{FR-Net} predicts mass distribution and contact states to enable quadrupedal recovery on challenging terrain~\cite{frnet}. VIGOR distills terrain-aware recovery goals into a depth-conditioned policy~\cite{vigor}; UniReLo combines terrain-relative support guidance with motion priors~\cite{unirelo}. RecoverFormer predicts useful environmental contact regions~\cite{recoverformer}, while confined-space planning addresses geometric restrictions during traversal~\cite{confined}. Our setting couples obstruction and manipulation: the surface blocking the rise may itself be pushed, lifted, or tipped. We reward the clearance produced by these interactions without specifying a contact sequence or escape direction.

\section{Problem Formulation}\label{sec:problem}
We consider a fallen humanoid whose recovery is obstructed by one or more overhead objects. 
The goal is to create enough space to regain sustained standing. 
For an immovable obstacle, clearance must come from moving the robot; while for a movable obstacle, it may also come from displacing the object through contact. 
Recovery, therefore, depends on the configuration of both robot and obstacle, their available motions, and the support needed to rise.

We instantiate this task in simulation with two plate types, with unobstructed ground providing a control condition.
A \emph{guided plate} allows only vertical translation, where it can be lifted but cannot slide or tilt, making lateral repositioning important. 
A \emph{free plate} is unattached and can be pushed, slid or tipped aside. 
Testing on a real Unitree G1 with fixed furniture and multi-object arrangements extends our evaluation beyond these training instances onto hardware.

The policy $a_t=\pi_\theta(o_t)$ uses only proprioception:
\begin{equation}
o_t=[\omega_t^b,g_t^b,q_t-q^0,\dot q_t,a_{t-1}]\in\mathbb R^{93}.
\label{eq:policy_observation}
\end{equation}
Here $\omega_t^b$ is the base angular velocity; $g_t^b$ denotes the projected gravity; $q_t-q^0$ is the joint-position offsets from the nominal pose; and $\dot q_t$ represents joint velocities. 
$a_t\in\mathbb R^{29}$ specifies joint-position targets tracked by PD controllers.
Obstacle geometry and contacts are available for learning but are not actor inputs.
Success requires sustained standing (Sec.~\ref{sec:setup}), not complete separation from the obstacle. 
A robot may push a plate aside yet fail to rise, or stand while still touching it. 
Escape therefore provides intermediate learning feedback, while standing determines task completion.

\section{Method}\label{sec:method}
\begin{figure*}[t]
\centering\input{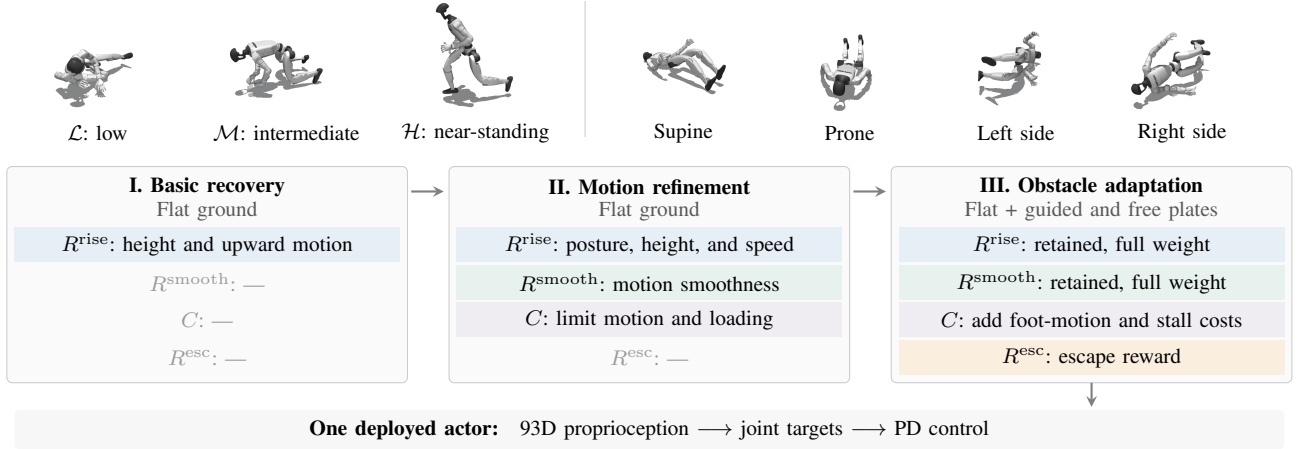}
\caption{Training overview. Top: example initial states from each $\mathcal{L}/\mathcal{M}/\mathcal{H}$ group and fallen-posture family. Bottom: reward groups active in each training stage, in the order and colors of Table~\ref{tab:reward_definitions}; dashes mark absent groups. The actor and its observation normalization carry over between stages, and a single actor is deployed.}
\label{fig:framework}
\end{figure*}

\subsection{Overview and Learning Objective}
\label{sec:overview}
We formulate humanoid recovery under overhead constraints as a partially observable Markov decision process (POMDP), optimized with PPO~\cite{ppo} and an asymmetric actor--critic. The actor maps the current 93-dimensional proprioceptive observation in Eq.~(\ref{eq:policy_observation}) to joint-position targets. The critic estimates return from ten 96-dimensional frames containing noise-free versions of the same observation components plus base linear velocity.

The objective combines a fixed motion prior, posture-dependent refinement, and contact-supported geometric guidance (Fig.~\ref{fig:framework}) to maximize expected discounted return while learning clearance creation and standing.

Learning proceeds in three stages. \emph{Stage~I (Basic)} discovers flat-ground recovery from a randomly initialized actor, using upward-motion and height objectives. \emph{Stage~II (Refine)} introduces posture-dependent recovery, settling, and execution costs. \emph{Stage~III (Adapt)} retains these objectives and adds obstacle interaction, geometric guidance, and direct foot-motion and joint-stall costs. Tables~\ref{tab:training_stages} and~\ref{tab:reward_definitions} summarize stage distributions and objectives.

The final-stage objective is expressed as the per-step reward
\begin{equation}
r_t=S_t\bigl(R_t^{\rm rise}+R_t^{\rm smooth}\bigr)-C_t+R_t^{\rm esc},
\label{eq:adaptation}
\end{equation}
where $S_t$ is the motion-prior score, $R^{\rm rise}$ encourages upright recovery, $R^{\rm smooth}$ encourages settling, $C\geq0$ collects execution costs, and $R^{\rm esc}$ guides clearance and escape. The product favors coordinated recovery; additive execution and escape terms preserve their feedback even when the prior score is low.

\begin{table}[t]
\centering\footnotesize
\caption{Initial-state distribution: low ($\mathcal{L}$), intermediate ($\mathcal{M}$), and near-standing ($\mathcal{H}$) percentages within each scene type.}
\label{tab:training_stages}
\setlength{\tabcolsep}{9.7pt}
\begin{tabular}{llrrr}
\toprule
Stage & Scene share & $\mathcal{L}$ & $\mathcal{M}$ & $\mathcal{H}$\\
\midrule
Stage I (Basic) & Flat: 100\% & 20 & 40 & 40\\
Stage II (Refine) & Flat: 100\% & 40 & 40 & 20\\
Stage III (Adapt) & Flat: 50\% & 40 & 40 & 20\\
 & Guided plate: 25\% & 100 & 0 & 0\\
 & Free plate: 25\% & 100 & 0 & 0\\
\bottomrule
\end{tabular}
\end{table}

\subsection{Prior-Guided Motion Refinement}
\label{sec:refinement}
We use the fixed SMP motion prior~\cite{smp}, whose score $S_t$ is computed from normalized diffusion prediction errors. Higher scores favor motion consistent with human get-up data. Normalization scales are estimated in Stage~I and frozen thereafter; refinement changes the recovery policy, not the prior model. The prior supplies a movement preference without requiring the robot to track a reference trajectory through an obstacle.

Within each recovery attempt, posture-dependent phases adjust the targets and execution costs as the robot progresses from lying to standing. Low postures allow the larger movements needed for repositioning and support formation. As the robot rises, motion and loading constraints become stronger, while the desired upward speed decreases near the target height. A continuous posture gate also increases penalties on foot and base motion near upright configurations, encouraging the robot to settle without restricting movement near the floor.

During obstacle adaptation, two additional costs discourage unstable support and sustained joint loading. The foot-motion cost $C^{\rm foot}$ penalizes foot movement near upright postures, encouraging the robot to settle after rising. The joint-stall cost $C^{\rm stall}$ penalizes prolonged high torque with little joint motion, discouraging persistent pushing against an obstacle without progress.

\begin{table*}[t]
\centering\footnotesize
\caption{Reward definitions and signed weights. Colors match Fig.~\ref{fig:framework}; dashes indicate inactive terms. Weights are multiplied by the 0.02\,s control period. \textit{Notation.} $[x]_+=\max(x,0)$, $\chi(x)=\operatorname{clip}(x,0,1)$, $\ell(x;a,b)=\chi((x-a)/(b-a))$, $\sigma(x)=x^2(3-2x)$. $t$: control step; $s$: recovery phase; $j$: joint; $\mathrm{on/full}$: ramp onset/saturation; $u_{\rm torso}=\chi(-g_z^b)$; $\overline{v_{\rm foot}^2}=\operatorname{mean}_{i\in\mathrm{feet}}\|v_i\|^2$. $\mathrm{target}$, $\mathrm{up/down}$, and $\mathrm{limit}$ denote posture targets, task speed bounds, and execution speed bounds; $\delta_h$: taper margin; $e_{\rm knee}$: knee-range violation; $k_s$: phase scales. $\Delta a_t=a_t-a_{t-1}$ and $\Delta^2a_t=\Delta a_t-\Delta a_{t-1}$. \textit{Loading.} $\rho_j=|\tau_j|/\tau_j^{\max}$; $T_j$: time above $\rho_{\rm on}$; $\dot q_{\rm stall}$: stall-speed threshold; $A_j$: recent joint excursion; $m_j^{\rm load}$: low-pass filtered $\rho_j^2$. $\mathcal J_{\rm high}$ selects the three largest effort costs; $\langle\cdot\rangle$ averages physics substeps. \textit{Escape gates.} Binary indicators select episodes containing a plate ($\delta^{\rm plate}$), plate states passing penetration, contact-force and initial-contact checks ($\delta_t^{\rm valid}$), prior plate contact in the episode ($\delta_t^{\rm contact}$), and currently confirmed escape ($\delta_t^{\rm escape}$). $g_t^{\rm progress}=\delta_t^{\rm valid}\delta_t^{\rm contact}(1-\delta_t^{\rm escape})$; $f_t^{\rm hand}$ is the fraction of hands providing support. $*$: terms summed in $R^{\rm geom}$; $\dagger$: scaled by $\min(n_{\rm update}/500,1)$ in Stage~II, fully active in Stage~III.}
\label{tab:reward_definitions}
\setlength{\tabcolsep}{4pt}\renewcommand{\arraystretch}{1.14}
\begin{tabular}{@{}p{2.95cm}p{10.0cm}rrr@{}}
\toprule
Term & Definition & I: Basic & II: Refine & III: Adapt\\
\midrule

\rowcolor{RiseBlue!12}[0pt][0pt]\multicolumn{5}{@{}l@{}}{\textbf{$R^{\rm rise}$: Recovery}}\\
Basic upward motion & $\begin{cases}e^{-100[0.25-\dot h_{\rm head}]_+^2},&h_{\rm head}<0.9\\1,&h_{\rm head}\geq0.9\end{cases}$ & 0.70 & -- & --\\
Basic height & $e^{-[1.1-h_{\rm head}]_+^2}$ & 0.30 & -- & --\\
Phase-dependent posture & $e^{-8[h_{{\rm head},s}^{\rm target}-h_{\rm head}]_+^2-6[u_{{\rm torso},s}^{\rm target}-u_{\rm torso}]_+^2-5e_{\rm knee}}$ & -- & 0.22 & 0.22\\
Head velocity & $e^{-45(\dot h_{\rm head}-v_{{\rm head},s}^{*})^2}\,e^{-140([\dot h_{\rm head}-v_{{\rm head},s}^{\rm up}]_+^2+[-\dot h_{\rm head}-v_{{\rm head},s}^{\rm down}]_+^2)}$\newline $v_{{\rm head},s}^{*}=\bar v_{{\rm head},s}\chi((h_{{\rm head},s}^{\rm target}-h_{\rm head})/\delta_h)$ & -- & 0.18 & 0.18\\
Head height & $e^{-2[1.15-h_{\rm head}]_+^2}$ & -- & 0.10 & 0.10\\
Torso uprightness & $\bigl(u_{\rm torso}\bigr)^2$ & -- & 0.15 & 0.15\\
\rowcolor{SettleGreen!12}[0pt][0pt]\multicolumn{5}{@{}l@{}}{\textbf{$R^{\rm smooth}$: Motion smoothness}}\\
Quiet feet & $r_{\rm quiet\,feet}=1-g_{\rm upright}+g_{\rm upright} e^{-20\overline{v_{\rm foot}^2}}$\newline $g_{\rm upright}=\sigma(\ell(h_{\rm head};h_{\rm on},h_{\rm full}))\,\sigma(\ell(u_{\rm torso};u_{\rm on},u_{\rm full}))$ & -- & 0.08 & 0.08\\
Quiet base & $1-g_{\rm upright}+g_{\rm upright} e^{-8\|v_{\rm base}\|^2}$ & -- & 0.07 & 0.07\\
Angular quietness & $e^{-0.8\|\omega_{\rm base}\|^2}$ & -- & 0.07 & 0.07\\
Joint quietness & $e^{-0.04\operatorname{mean}_j\dot q_j^2}$ & -- & 0.06 & 0.06\\
Action smoothness & $e^{-4\operatorname{mean}_j(\Delta a_j)^2}$ & -- & 0.07 & 0.07\\
\rowcolor{LoadPurple!12}[0pt][0pt]\multicolumn{5}{@{}l@{}}{\textbf{$C$: Execution costs}}\\
Action rate$^\dagger$ & $k_s^{\Delta a}\sum_j(\Delta a_j)^2$ & -- & $-0.0015$ & $-0.0015$\\
Action acceleration$^\dagger$ & $k_s^{\Delta^2a}\sum_j(\Delta^2a_j)^2$ & -- & $-0.0012$ & $-0.0012$\\
Joint acceleration$^\dagger$ & $\langle k_s^{\ddot q}\sum_j\ddot q_j^2\rangle$ & -- & $-5\!\times\!10^{-8}$ & $-5\!\times\!10^{-8}$\\
Torque$^\dagger$ & $\langle k_s^\tau\sum_j\tau_j^2\rangle$ & -- & $-10^{-6}$ & $-10^{-6}$\\
Joint overspeed$^\dagger$ & $\langle\sum_j[|\dot q_j|-\dot q_s^{\max}]_+^2\rangle$ & -- & $-0.02$ & $-0.02$\\
Joint overpower$^\dagger$ & $\langle\operatorname{mean}_j[|\tau_j\dot q_j|-P_s^{\max}]_+^2\rangle$ & -- & $-2\!\times\!10^{-6}$ & $-2\!\times\!10^{-6}$\\
Head overspeed$^\dagger$ & $\langle[\dot h_{\rm head}-v_{{\rm head},s}^{\rm limit}]_+^2+[-\dot h_{\rm head}-0.20]_+^2\rangle$ & -- & $-1$ & $-1$\\
Sustained effort$^\dagger$ & $\left\langle\frac{1}{3}\sum_{j\in\mathcal J_{\rm high}}c_j^{\rm effort}\right\rangle$\newline $c_j^{\rm effort}=([\sqrt{m_j^{\rm load}}-0.45]_+/0.55)^2$ & -- & $-0.05$ & $-0.05$\\
Soft joint limits & $\sum_j([q_j^{\min}-q_j]_++[q_j-q_j^{\max}]_+)$ & -- & $-0.10$ & $-0.10$\\
Foot motion & $C_t^{\rm foot}=\ell(h_{\rm head};h_{\rm on},h_{\rm full})\ell(u_{\rm torso};u_{\rm on},u_{\rm full})\min(\overline{v_{\rm foot}^2}/0.1^2,10)$ & -- & -- & $-0.03$\\
Joint stall & $C_t^{\rm stall}=\max_j\{\langle c_j^{\rm stall}\rangle\chi(1-A_j/0.10)\}$\newline $c_j^{\rm stall}=\ell(\rho_j;\rho_{\rm on},\rho_{\rm full})^2\ell(T_j;T_{\rm on},T_{\rm full})\chi(1-|\dot q_j|/\dot q_{\rm stall})$ & -- & -- & $-0.20$\\
\rowcolor{SupportOrange!14}[0pt][0pt]\multicolumn{5}{@{}l@{}}{\textbf{$R^{\rm esc}$: Escape reward}}\\
Supported progress$^*$ & $r_t^{\rm progress}$ in (\ref{eq:guidance}): best-so-far depth/gap improvement & -- & -- & 0.45\\
Dense clearance$^*$ & $\delta_t^{\rm valid}\{0.85\chi(1-N_t/\max(N_0,1))+0.15\chi(c_t/0.04)\}$ & -- & -- & 0.08\\
Root--plate separation$^*$ & $\delta_t^{\rm contact}\chi\big((d_{\mathrm{root,plate}}(t)-d_{\mathrm{root,plate}}^{\max}(t))/0.025\big)$ & -- & -- & 0.01\\
Escape completion & $\delta_t^{\rm escape}$ & -- & -- & 0.60\\
Excess plate force & $\delta^{\rm plate}([F_t-300]_+/300)^2$ & -- & -- & $-0.03$\\
\bottomrule
\end{tabular}

\end{table*}

\subsection{Contact-Supported Geometric Guidance}
\label{sec:geometric_reward}
Our guidance measures how contact and relative motion create space for recovery. It must recognize progress both when the robot moves out and when the obstacle is displaced, without selecting an escape direction in advance. We capture this interaction through supported progress, dense clearance, and root--plate separation. Their weighted sum $R^{\rm geom}$ enters the escape objective $R^{\rm esc}$ alongside completion and excess-force terms.

\begin{figure*}[t]
\centering\includegraphics[width=\textwidth]{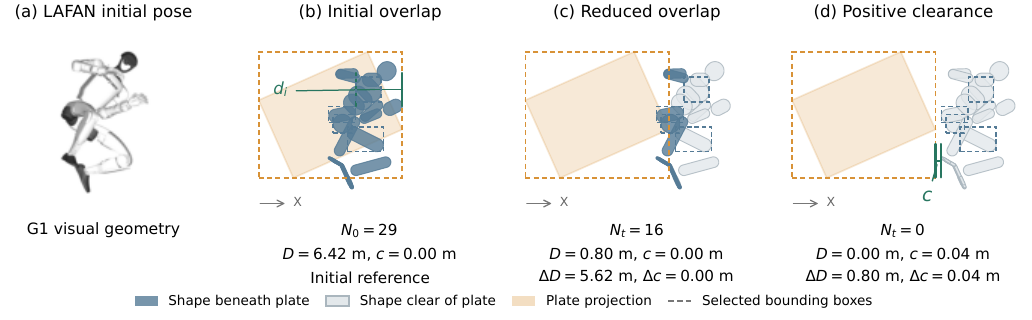}
\caption{Geometric measures for a supine LAFAN1 pose. (a) G1 geometry. (b--d) The same pose translated relative to a yawed, tilted plate, with selected world-axis-aligned bounds. $N$ counts the collision shapes beneath the plate (blue), $d_i$ is the depth of one shape, and $D$ sums the depths over all 29 shapes. $N$ and $D$ decrease before a positive gap $c$ appears. The configurations are constructed for illustration and are not taken from a rollout.}
\label{fig:escape}
\end{figure*}

We project the world-axis-aligned bounds of the robot's 29 collision shapes and the plate onto the ground plane (Fig.~\ref{fig:escape}). For an overlapping robot box $i$, let $o_{ix},o_{iy}$ be the translations along each horizontal axis needed to leave the plate box. Its covered depth is
\begin{equation}
d_i=\min(o_{ix},o_{iy}),\qquad D_t=\sum_i d_i,
\end{equation}
with $d_i=0$ for non-overlapping boxes. The clearance gap $c_t$ is the minimum horizontal distance between any body box and the plate box. Thus $D_t$ measures progress while the body is partly covered, whereas $c_t>0$ only after all projected boxes separate. Individual shapes register a limb clearing before the whole body exits. Covered depth is a distance, not an intersection area: it reflects how far an individual projected shape must move to clear a plate edge. A limb can therefore reduce $D_t$ even while the covered-shape count remains unchanged. These planar bounds approximate rotated geometry and ignore vertical ordering; the escape criterion separately excludes body parts entirely above the plate.

Let $\Delta D_t$ be the improvement below the episode's previous minimum depth, and $\Delta c_t$ the improvement above its previous maximum gap. Rewarding these new records prevents repeated payment for revisiting the same clearance:
\begin{equation}
\begin{split}
r_t^{\rm progress}={}&g_t^{\rm progress}f_t^{\rm hand}\mathbf1[h_{{\rm head},t}\leq 0.9]\\
&\times\big[\chi(\Delta D_t/0.025)+\tfrac12\chi(\Delta c_t/0.02)\big].
\end{split}
\label{eq:guidance}
\end{equation}
Distances are in meters. The gate $g_t^{\rm progress}$ requires valid plate states, prior contact, and unconfirmed escape; $f_t^{\rm hand}$ measures hand support (Table~\ref{tab:reward_definitions}). This rewards supported low-posture progress without prescribing a direction. Prior contact need not persist: progress can continue as contact is released. Confirmed escape switches off this term, while the standing objective remains active.

Dense clearance rewards the current fraction of shapes no longer beneath the plate and the current gap; $N_t,N_0$ are the current and initial covered-shape counts. Root--plate separation rewards increases beyond the previous maximum horizontal root-to-plate-center distance. Escape completion rewards confirmed clearance with no plate contact for 0.3\,s, and is revoked upon return. Finally, the force cost penalizes the largest plate-contact force norm $F_t$ above 300\,N.

Supported progress rewards newly created space, whereas dense clearance rewards maintaining it. Both provide feedback before recovery succeeds; the standing objective still requires the robot to organize support and rise.

\subsection{Training Curriculum and Implementation}
\label{sec:training_details}
\label{sec:curriculum}
To practice the whole recovery, we sample initial states from retargeted LAFAN1 get-up clips~\cite{lafan}, grouped by height and uprightness into low ($\mathcal{L}$), intermediate ($\mathcal{M}$), and near-standing ($\mathcal{H}$) states. These groups determine where practice starts; they do not prescribe trajectories. Within $\mathcal{L}$, 75\% of states come from clips and 25\% from perturbed procedural supine, prone, left-side, and right-side poses. Other groups use clips only. Sampling is stratified by posture; invalid states are rejected and velocities reset to zero. Intermediate and near-standing starts let the policy practice support transfer before it can reliably reach these states from the floor; procedural low poses ensure coverage of each fallen-posture family.

Table~\ref{tab:training_stages} allocates these groups across stages. Stage~II increases low starts; Stage~III assigns half the environments to flat ground and one quarter each to guided and free plates, with all obstacle episodes starting low. The four fallen-posture families are equally likely in Stages~I and~III; Stage~II assigns 50\% to prone and divides the remainder equally.

Within an episode, the reward phase $s$ progresses through low, transition, rising, and standing. Advancement requires height, uprightness, knee-angle, and vertical-speed conditions to hold for 0.2\,s; standing additionally requires both-foot support for 0.5\,s. A fall returns the phase to low. These phases are distinct from the initial-state groups and are not actor inputs.

The actor outputs targets at 50\,Hz, tracked by PD controllers with MuJoCo~\cite{mujoco} dynamics at 500\,Hz.

We train 4096 parallel environments, transferring the actor and observation normalization between stages while reinitializing the critic and optimizer. Stages~II--III disable standing and low-prior early termination, allowing continued practice after rising or a failed attempt. Randomization covers observation noise, external pushes, friction, encoder bias, center of mass, link masses and inertias, motor gains, and command delay. Plate mass is randomized; dimensions remain fixed during training.

\section{Experiments}
Our experiments ask three questions: do open-space recovery skills transfer to overhead constraints, which reward components improve recovery and execution, and how well does single-plate training transfer to physical obstacles?

\subsection{Experimental Setup}
\label{sec:setup}
\emph{Conditions.} We evaluate three simulated conditions. \emph{Flat} is unobstructed ground. \emph{Guided} uses a $0.90\times0.64\times0.07$\,m, 6\,kg plate that can be lifted but cannot translate sideways or rotate, making lateral movement important for creating room to rise. \emph{Free-R} uses a free plate whose length and width are each scaled by a factor in $[1,1.75]$, with mass in $[6,18]$\,kg, horizontal offset within $\pm0.15$\,m and yaw within $\pm45^\circ$; training uses only the nominal size, so larger plates are unseen. Every trial starts from a low ($\mathcal{L}$) state, with supine, prone, left-side and right-side starts equally likely and 25\% of states drawn from the procedural fallen-posture families. Each obstacle condition contains 512 trials and Flat contains 1024. All methods share the same initial states and Free-R draws. Evaluation disables external pushes and observation noise.

\emph{Metrics.} A trial succeeds (SR) if the robot reaches standing within 20\,s and then remains standing for 3\,s. Standing requires a head height of at least 1.15\,m, base uprightness of at least 0.90, and a vertical ground reaction above 20\,N on each foot. We also measure execution quality over the full 23\,s window, including failed trials. \emph{Slip} is the travel of contact feet per second while the robot is near upright (head above 0.78\,m, uprightness above 0.70). HLS$_{95}$ is the 95th percentile, across trials, of the time the most loaded joint spends above 70\% of its torque limit while moving slower than 0.3\,rad/s.

\emph{Policies.} For each training configuration, we report the checkpoint with the highest combined success under the guided plate and a fixed-size free plate among its training runs.

\subsection{Comparison with Recovery Methods}
We compare with HoST~\cite{host} and HumanUP~\cite{humanup}, two learning-based get-up methods for humanoids, and with an open-source G1 controller that learns locomotion and fall recovery in a single policy using an adversarial motion prior~\cite{amp,ampmjlab}. All three use their original training configurations, which contain no overhead obstacles, and are evaluated in our conditions without retraining. This comparison tests whether recovery skills learned in open space carry over to overhead constraints.
\begin{table}[t]
\centering\footnotesize
\caption{Comparison with humanoid recovery methods (SR, \%~$\uparrow$). $\pm$ denotes the 95\% confidence half-width (normal approximation) over 1024 flat and 512 plate trials with shared initial states. $^*$1024/1024 successes; Wilson 95\% interval [99.6, 100].}
\label{tab:main_comparison}
\setlength{\tabcolsep}{4pt}\renewcommand{\arraystretch}{1.12}
\begin{tabular*}{\columnwidth}{@{\extracolsep{\fill}}lccc@{}}
\toprule
Method & Flat & Guided & Free-R\\
\midrule
HoST~\cite{host}       & 98.7$\pm$0.7 & 59.8$\pm$4.2 & 80.9$\pm$3.4\\
HumanUP~\cite{humanup} & 97.6$\pm$0.9 & 34.4$\pm$4.1 & 74.4$\pm$3.8\\
AMP~\cite{ampmjlab}    & 68.1$\pm$2.9 & 2.1$\pm$1.3 & 4.9$\pm$1.9\\
Ours                   & \textbf{100.0}$^*$ & \textbf{88.5$\pm$2.8} & \textbf{98.6$\pm$1.0}\\
\bottomrule
\end{tabular*}

\end{table}

The comparison separates the ability to stand from the ability to cope with obstruction (Table~\ref{tab:main_comparison}). HoST and HumanUP succeed on nearly all flat trials, but lose success beneath plates. AMP also struggles on flat ground, suggesting a mismatch between its training distribution and our fallen states. Guided is particularly challenging because the plate cannot be pushed aside; in Free-R, upward movement can instead displace the obstacle. Our policy attains the highest success in both conditions, consistent with the benefit of practicing clearance creation alongside standing.

This comparison concerns transfer of the released recovery skills, not the relative capacity of their learning algorithms after matched obstacle training. The baseline policies have not practiced overhead interaction, while ours has. The reward ablations below provide the more direct test of our design: they remove geometric feedback or an execution cost within the obstacle-training setting. Together, the comparisons distinguish the challenge posed by obstruction from the contribution of the proposed objective.

\begin{figure*}[t]
\centering\input{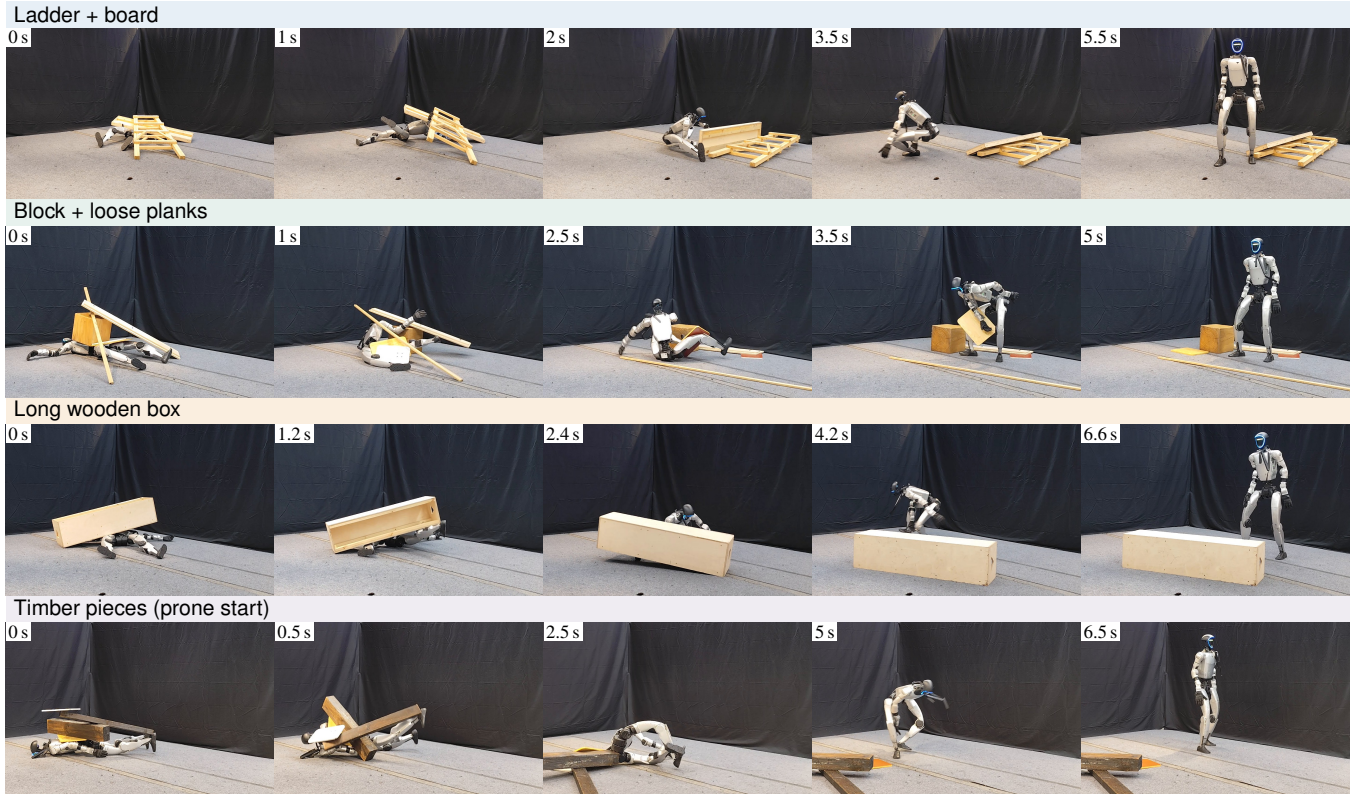}
\caption{Hardware recovery among obstacle arrangements absent from training: a ladder and board, a block with loose planks, a long wooden box, and timber pieces. The first three rows start supine and the bottom row starts prone. Each row shows frames from one trial; timestamps are relative to the first frame shown.}
\label{fig:hardware_transfer}
\end{figure*}
\subsection{Ablation of the Training Objective}
Table~\ref{tab:ablation} removes one reward component at a time and keeps the other terms unchanged. Each row reports that configuration's selected checkpoint.
\begin{table}[t]
\centering\footnotesize
\caption{Ablation of the training objective. Slip and HLS$_{95}$ pool both obstacle conditions. $\pm$ denotes the 95\% confidence half-width over evaluation trials (normal approximation for SR; bootstrap for Slip and HLS$_{95}$). Checkpoint selection follows Sec.~\ref{sec:setup}.}
\label{tab:ablation}
\setlength{\tabcolsep}{3pt}\renewcommand{\arraystretch}{1.12}
\begin{tabular*}{\columnwidth}{@{\extracolsep{\fill}}lcccc@{}}
\toprule
 & \multicolumn{2}{c}{SR (\%) $\uparrow$} & Slip $\downarrow$ & HLS$_{95}$ $\downarrow$\\
\cmidrule(lr){2-3}
Configuration & Guided & Free-R & (mm/s) & (s)\\
\midrule
Ours                       & \textbf{88.5$\pm$2.8} & \textbf{98.6$\pm$1.0} & \textbf{5.8$\pm$0.3} & \textbf{16.0$\pm$1.0}\\
w/o $R^{\rm geom}$         & 18.9$\pm$3.4 & 83.0$\pm$3.3 & 8.6$\pm$0.5 & 18.7$\pm$1.7\\
w/o foot-motion cost $C^{\rm foot}$   & 83.2$\pm$3.2 & 93.8$\pm$2.1 & 19.2$\pm$0.5 & 19.3$\pm$0.5\\
w/o joint-stall cost $C^{\rm stall}$   & 84.6$\pm$3.1 & 97.3$\pm$1.4 & 12.6$\pm$0.3 & 21.3$\pm$0.3\\
\bottomrule
\end{tabular*}

\end{table}

\emph{Creating room to rise.} Removing $R^{\rm geom}$ leaves escape completion and the excess-force cost, but eliminates feedback during clearance creation. Guided success falls from 88.5\% to 18.9\%, and Free-R success from 98.6\% to 83.0\%. The larger Guided loss is consistent with the difficulty of discovering lateral escape from a completion reward alone. Geometric feedback also accommodates movable obstacles: coverage decreases whether the robot leaves the plate or pushes it away.

\emph{Controlling the rise.} The two execution costs address behavior beyond escape. Removing either retains substantial Guided success but increases slip and high-load stall duration. These metrics directly measure the targeted behaviors; the accompanying changes in other outcomes are also informative. Without $C^{\rm stall}$, slip more than doubles; without $C^{\rm foot}$, Free-R success decreases to 93.8\%. Thus both costs contribute to recovery quality, rather than simply maximizing the standing count. As a separate sensitivity test, increasing upper-body mass and inertia by 30\% without retraining reduces Guided success from 88.5\% to 85.5\%.

\subsection{Hardware Experiments}
We deploy a policy trained with the complete objective on the Unitree G1 and test it in eight scenes (Table~\ref{tab:hardware_success}): flat ground as a control, four single obstacles, and three arrangements of multiple objects. Apart from the single board, these obstacles differ from the training plates in shape, mass and constraint. In each attempt, the robot is placed beneath the obstacle with a randomly chosen heading, and attempts within a scene are run consecutively. Hardware SR uses the same 20\,s recovery deadline and 3\,s standing confirmation, observed through 23\,s. The trials in Figs.~\ref{fig:teaser} and~\ref{fig:hardware_transfer} use the same policy.
\begin{table}[t]
\centering\footnotesize
\caption{Hardware recovery success across obstacle arrangements. Attempt counts differ by scene. The bottom row excludes the flat-ground control. Intervals are Wilson 95\% confidence intervals.}
\label{tab:hardware_success}
\begingroup
\setlength{\tabcolsep}{4pt}\renewcommand{\arraystretch}{1.12}
\begin{tabular}{@{}lccc@{}}
\toprule
Scene & Successes/$N$ & SR (\%) $\uparrow$ & 95\% CI\\
\midrule
Flat ground (control) & 20/20 & 100.0 & [83.9, 100.0]\\
\midrule
\multicolumn{4}{@{}l}{\emph{Single obstacle}}\\
\quad Single movable board & 10/13 & 76.9 & [49.7, 91.8]\\
\quad Under table & 27/35 & 77.1 & [61.0, 87.9]\\
\quad Rigid equipment case & 12/16 & 75.0 & [50.5, 89.8]\\
\quad Long wooden box & 12/14 & 85.7 & [60.1, 96.0]\\
\midrule
\multicolumn{4}{@{}l}{\emph{Multiple objects}}\\
\quad Board + ladder & 7/29 & 24.1 & [12.2, 42.1]\\
\quad Block + loose planks & 8/10 & 80.0 & [49.0, 94.3]\\
\quad Timber arrangement & 19/33 & 57.6 & [40.8, 72.8]\\
\midrule
All obstacle scenes & 95/150 & 63.3 & [55.4, 70.6]\\
\bottomrule
\end{tabular}

\endgroup
\end{table}

The policy stands in every flat-ground attempt and in 95 of 150 obstacle attempts (Table~\ref{tab:hardware_success}). Obstacle mobility appears to influence the observed recovery behavior. A board, a case or a long box can be pushed or tipped aside, and the robot then rises as it does under a free plate, though less reliably than in simulation. The table cannot be lifted or slid, and the illustrated recovery proceeds by rolling out past an edge (Fig.~\ref{fig:teaser}); unlike the guided plate, it cannot even be raised. Arrangements of several objects vary. Where the pieces leave an exit, as with loose planks on a block, the policy often gets out, although this scene has only ten attempts. Where a limb can be caught and held, as with a ladder under a board, it usually fails.

\subsection{Failure Analysis}
\label{sec:failures}
We observed board--ladder failures involving mechanical entanglement. As the robot pushes up beneath the board, an arm can pass between the ladder rungs. During the subsequent rise, a rung hooks the forearm, and the robot either lifts the ladder with it or stalls with the shoulder and elbow restrained. The timber arrangement produces similar trapping when a limb wedges between pieces. Beneath the table, the robot can catch on a corner and remain stuck short of the edge. Single-plate training does not cover trapping by ladder rungs or gaps between multiple objects. Proprioception does not explicitly represent this contact geometry.
\section{Discussion and Conclusion}
We presented a learning approach for humanoid recovery beneath overhead obstacles. A geometric reward encourages the robot to create room to rise, while a motion prior and posture-dependent costs shape the motion. A single proprioceptive policy transfers to physical scenes that differ from the single-plate training distribution, including furniture and multi-object arrangements.

\emph{Limitations.} The current policy does not reliably resolve limb entanglement in the tested scenes (Sec.~\ref{sec:failures}). Training uses a single plate, and proprioceptive observations provide no explicit representation of surrounding contact geometry. Covered depth and the clearance gap approximate tilted geometry with axis-aligned bounds. Training with multiple interacting objects and adding exteroceptive or tactile feedback are natural next steps.

\bibliographystyle{IEEEtran}
\bibliography{paper_references}
\end{document}